# An Investigation of Quantum Deep Clustering Framework with Quantum Deep SVM & Convolutional Neural Network Feature Extractor

ARIT KUMAR BISHWAS, Amity Institute of Information Technology, Noida, India
ASHISH MANI, Department of EEE, Amity University, Noida, India
VASILE PALADE, Faculty of Engineering and Computing Coventry University, Coventry, UK

In this paper, we have proposed a deep quantum SVM formulation, and further demonstrated a quantum-clustering framework based on the quantum deep SVM formulation, deep convolutional neural networks, and quantum $K$-Means clustering. We have investigated the run time computational complexity of the proposed quantum deep clustering framework and compared with the possible classical implementation. Our investigation shows that the proposed quantum version of deep clustering formulation demonstrates a significant performance gain (exponential speed up gains in many sections) against the possible classical implementation. The proposed theoretical quantum deep clustering framework is also interesting & novel research towards the quantum-classical machine learning formulation to articulate the maximum performance.

## 1 INTRODUCTION

Clustering is a very interesting and complex problem. It has a vast range of applications but solving the problem for non-linear big data is a tremendous challenge. There are many interesting clustering algorithms have been developed **[1]** over time**,** but are not good at handling non-linear big data sets. Many research groups have demonstrated that supervised approach helps in handling a clustering problem in a more effective way **[2]**. In recent development **[3]**, a research group from Facebook has demonstrated a very interesting classical deep clustering framework using a pre-trained deep convolutional neural network (CNN) **[4]** and classical $K$-Means clustering **[5]** algorithm. In this paper, we propose an improvising formulation of the deep clustering with quantum mechanical postulates. Our proposed quantum deep clustering framework demonstrates a very significant performance gain against possible classical approaches. Our proposed framework also set up one of the first of its kind novel formulation for a classical-quantum fusion machine learning framework.

In our research, we first develop a possible deep quantum SVM formulation. We then, augment this quantum deep SVM formulation to design a quantum deep clustering framework. We take the help of a classical pre-trained deep CNN to extract the features of the unlabelled non-linear input big data. These extracted features are then used to train a quantum deep SVM. Quantum $K$-Means clustering algorithm also uses these extracted features to generate optimized clusters. These optimal clusters then help in generating pseudo labels for the non-linear input big data.

Quantum deep SVM uses these pseudo labels (in association with the non-linear input big data) for solving its cost function. Then based on the quantum classifier's (in this case quantum deep SVM) loss calculation, we retrain the deep CNN using back-propagating the gradients w.r.t. the CNN weights parameters. This process keeps on iterating until we reach to acceptable optimized cluster assignments by quantum $K$-Means clustering algorithm **[6]**.

## 2    SUPPORT VECTOR MACHINE

With given $M$ training data points of the form $\{(\vec{x}_i, y_i); x \in \mathbb{R}^N, y = \pm 1\}_{i=1,2,...M}$, where $y_i$ represents the binary class values (either $+1$ or $-1$) to which $\vec{x}_i$ belongs to, the very interesting task of a support vector machine is to classify a vector $\vec{x}$ into two classes. The SVM define a decision boundary $w^T \phi(\vec{x}) + b = 0$, , and find a maximum marginal (optimal) hyperplane around the decision boundary to classify the binary data points. These hyperplanes are constructed as:

$$w^T \phi(\vec{x}_i) + b > 0, \text{ for } \vec{x}_i \text{ to be in class } +1, \text{ and} \tag{1}$$

$$w^T \phi(\vec{x}_i) + b < 0, \text{ for } \vec{x}_i \text{ to be in class } -1. \tag{2}$$

A maximum margin is defined by the maximum distance,

$$\frac{2|\vec{w}^T \phi(\vec{x}_s) + b|}{\|\vec{w}\|_2} \approx \frac{2}{\|\vec{w}\|_2}, \text{ where } \vec{x}_s \text{ is the support vector.} \tag{3}$$

between the two parallel hyperplanes as shown in **Fig.1**.

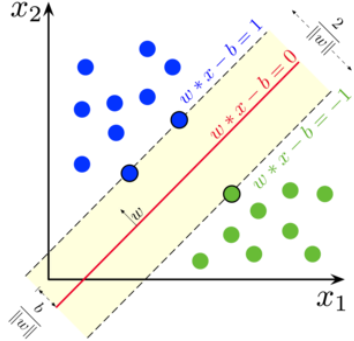

*Fig.1: The hyperplanes formulation in support vector machine. The picture has been referred from Wikipedia [7])*

The goal here is to optimize the following function

$$arg\ max_w \frac{1}{\|\vec{w}\|_2}; s.t\ min_i\ y_i(w^T \phi(\vec{x}_i) + b) = 1 \tag{4}$$

$$\approx min_w \frac{1}{2}\|\vec{w}\|_2; s.t\ y_i(w^T \phi(\vec{x}_i) + b) \geq 1, \forall i \tag{5}$$

The above equation is known as the primal form of SVM.

Till now, in its simplest linear form, we assumed that there are no data points inside the margin. But this ideal case is not possible with real-life data points where some data points will be inside the margin. To handle this situation, we introduce a penalty $C = \begin{cases} 0: less\ complex\ boundary \\ \infty: more\ complex\ boundary \end{cases}$, and a non-negative slack variable $\xi$. The optimization function then will be

$$min_w \frac{1}{2}\|\vec{w}\|_2 + C\sum_i \xi_i \quad (6)$$

$$s.t\ y_i(w^T\phi(\vec{x}_i) + b) \geq 1 - \xi_i\ \forall i, \xi_i \geq 0\ \forall i$$

In the least square SVM formulation, the optimization function is slightly different from the above equation and defines as,

$$min_w \frac{1}{2}\|\vec{w}\|_2 + \frac{\zeta}{2}\sum_{i=1}^M e^2_i \quad (7)$$

$$s.t\ y_i(w^T\phi(\vec{x}_i) + b) \geq 1 - e_i\ \forall i, e_i \geq 0\ \forall i,$$

where $e_i$ is the slack value. We then construct a Lagrange function to solve the above least square SVM optimization function.

$$L(w, b, e, \alpha) = \frac{1}{2}w^T w + \frac{\eta}{2}\sum_{i=1}^M e_i^2 - \sum_{i=1}^M \alpha_i\{[w^T\phi(\vec{x}_i) + b] + e_i - y_i\}, \quad (8)$$

where $\zeta$ and $\mu$ are hyper parameters for tuning the amount of regularization versus the sum squared error, and $\eta = \frac{\zeta}{\mu}$. $\alpha_i \in \mathbb{R}$ is the Lagrange multiplier, which is on the role of the distance from the margin and not sparse in general. After taking the partial derivatives of the equation **(8)** w.r.t. , $b, e_i\ and\ \alpha_i$, and eliminating the variables $w$ and $e_i$, we get a least-squares approximation form of the SVM:

$$\begin{bmatrix} 0 & \vec{1}^T \\ \vec{1} & K + \eta^{-1}I \end{bmatrix}\begin{bmatrix} b \\ \vec{\alpha} \end{bmatrix} = \begin{bmatrix} 0 \\ \vec{y}_i \end{bmatrix} = F\begin{bmatrix} b \\ \vec{\alpha} \end{bmatrix}, \quad (9)$$

with $K \in \mathbb{R}^{M \times M}$ is a kernel matrix which is defined as $K(\vec{x}_s, \vec{x}_i) = [\phi(\vec{x}_s)^T, \phi(\vec{x}_i)]$, $\vec{x}_s$ is the support vector, $\vec{y}_i = (y_1, y_2, \ldots, y_M)^T$, $\vec{1} = (1,1,\ldots,1)^T$, $I$ is $M \times M$ identity matrix, and $\vec{\alpha} = (\alpha_1, \alpha_2, \ldots, \alpha_M)^T$.

The SVM parameters are then determined by

$$[b, \vec{\alpha}]^T = \begin{bmatrix} 0 & \vec{1}^T \\ \vec{1} & K + \eta^{-1}I \end{bmatrix}^{-1} \begin{bmatrix} 0 \\ \vec{y}_i \end{bmatrix} = F^{-1}\begin{bmatrix} 0 \\ \vec{y}_i \end{bmatrix} \quad (10)$$

3   QUANTUM SUPPORT VECTOR MACHINE

For the quantum version of least square support vector machine, by describing the hyperplane with quantum matrix inversion algorithm **[8],** we generate a quantum state $|b, \vec{\alpha}\rangle$ **[9]**. For classifying a state $|\vec{x}\rangle$, we use a swap test between $|b, \vec{\alpha}\rangle$ and a query state $|\vec{x}\rangle$, and then the success probability determines the classification. In the quantum setting, we solve the following equation to determine the SVM parameters:

$$|b, \vec{\alpha}\rangle = \hat{F}^{-1}|\vec{y}\rangle \tag{11}$$

We now split $\hat{F}$ as $\hat{F} = (K + J + \eta^{-1}I)/trF$, so that we can apply the matrix inversion algorithm which helps to simulate the matrix exponential of $\hat{F}$. Where $J$ is star graph and defined as $J = \begin{bmatrix} 0 & \vec{1}^T \\ \vec{1} & 0 \end{bmatrix}$, and $\hat{F} = F/trF$ is the normalized version of $F$, $trF$ is the trace of $F$. By using the Lie product formulation, we get the exponential as

$$e^{-i\hat{F}\Delta t} = e^{-iK\Delta t/trF} e^{-iJ\Delta t/trF} e^{-i\eta^{-1}I\Delta t/trF} + O(\Delta t^2) \tag{12}$$

Here, with the reference **[10]**, the exponentials of sparse matrix $J$ and constant multipliers are easy to simulate. The $K$ is not sparse, so we determine the exponential of $K$ with quantum self-analysis technique. As per the discussion in **[10]**, with given multiple copies of a density matrix $P$, we can perform $e^{-iPt}$. In quantum self-analysis, a state functions as a Hamiltonian after got exponentiation and actively contributes to its quantum measurement. We get the normalized kernel $\hat{K} = K/trK = 1/trK \sum_{i,j=1}^{M} \langle \vec{x}_i | \vec{x}_j \rangle |\vec{x}_i||\vec{x}_j||i\rangle\langle j|$. By using the strategy developed in **[10]**, we do the exponentiation of $\hat{K}$ as:

$$e^{-i\mathcal{L}_{\hat{K}}\Delta t}(I) \approx I - i\Delta t[\hat{K}, I] + O(\Delta t^2), \text{ where } \mathcal{L}_{\hat{K}} = [\hat{K}, I] \tag{13}$$

Using equation **(12)**, we do phase estimation to determine the eigenvalues and corresponding eigenvectors. We then express $|\vec{y}\rangle$ in the form of eigenvectors, and invert the eigenvalues to achieve the required solution $|b, \vec{\alpha}\rangle$. This result,

$$|b, \vec{\alpha}\rangle = \frac{1}{\sqrt{b^2 + \sum_{i=1}^{M} \alpha_i^2}} (b|0\rangle + \sum_{i=1}^{M} \alpha_i |i\rangle) \tag{14}$$

We extend our discussion on the multiclass case now. As we know that the SVM is a binary class classification algorithm, but we can extend the binary SVM to multiclass SVM. In **[11][12]**, we have developed and discussed two different quantum multiclass SVM formulations. Here we explore the quantum multiclass SVM with quantum all-pair approach **[12]** in brief. In this approach, we first develop $g(g-1)/2$ quantum binary classifiers, where $g$ is the number of classes. Then, all the $g(g-1)/2$ quantum binary classifiers classify a quantum query state $|\vec{x}\rangle$. Using the quantum all-pair approach the desired class is predicted. We define the quantum all-pair algorithm as follows:

**ALGORITHM 1:** *Quantum All-Pair Algorithm*

1: *initialize class = any random element*, $1 \leq class\_index \leq g$
2: *initialize* $|V_q\rangle$ *as the vector of all classified classes*
3: *initialize frequency estimate* $s_{class\_index}$ *with any very small value*
4: *INITIAL-FREQUENCY-COUNT (* $class\_index, s_{class_{index}}$ *)*
5: *while* $(total\ running\ time\ < O\ (log g))$
6:      *initialize the memory* $|C_i\rangle = \sum_j \beta_{i,j} |j\rangle$

7:       *initialize the memory as*
          $|\psi_c\rangle = \sum_{i=0}^{g-1}|i\rangle|C_i\rangle|class\_index\rangle|s_{class\_index}\rangle$
8:       GROVER-QUANTUM-SEARCH ( $|\psi_c\rangle, |V_q\rangle, class\_index, s_{class\_index}$)
9:       MEASURE-REGISTER ($|\psi_c\rangle$)
10:      *if* $\left(s_{class\_index\_new} > s_{class\_index} + \frac{\varepsilon}{2g}\right)$
11:           $class\_index = class\_index_{new}$
12:           $s_{class\_index} = s_{class\_index\_new}$
13: *return* $|V_q\rangle[class\_index]$

Where *class_index* is a random index variable. The INITIAL-FREQUENCY-COUNT (*class_index*, $s_{class\_index}$) function imposts the initial frequency estimate $s_{class\_index}$ of *class_index*, $|C_i\rangle$ is a count memory state, which describes the number of times the $i^{th}$ class occurs in vector $|V_q\rangle$ defined as a fraction of $\left(\frac{g(g-1)}{2}\right)$. $|V_q\rangle$ holds the list of the predicted classes predicted by $\left(\frac{g(g-1)}{2}\right)$ quantum binary classifiers. The state $|C_i\rangle$ is in a superposition of all the values of $|V_q\rangle$. GROVER-QUANTUM-SEARCH ($|\psi_c\rangle, |V_q\rangle, class\_index, s_{class\_index}$) is Grover's quantum search on $|V_q\rangle$ for finding out the final class value. For each register of the state $|\psi_c\rangle$, the quantum measurement is prepared with the function MEASURE-REGISTER ($|\psi_c\rangle$). Based on the condition ($s_{class\_index\_new} > s_{class\_index} + \frac{\varepsilon}{2g}$), the *class_index* and $s_{class\_index}$ are updated, and $\varepsilon$ is the measurement error. The algorithm ensures the desired predicted class value $|V_q\rangle[class_{index}]$.

## 4 QUANTUM DEEP SVM FORMULATION

In recent years, deep learning demonstrated very promising results with neural networks**[13][14][15]**. In a similar way, few research groups have discussed deep learning architecture with SVMs in classical paradigm **[16][17]**. In deep SVM architecture, a deep neural network like architecture has been proposed, where the perceptrons have been replaced with an SVM.

The input training data is feed into the first layer of the network, where $z \subseteq v$ number of SVMs are present in the first layer $l^{(1)}$. Each of the $z$ SVMs in the first layer learns a single feature from the input data after being trained. The outputs of the $z$ SVMs in the first layer are the activation values and act as the input for the next layer. And, the process is kept on going till the last layers where there is multiclass SVM which predicts a final class value. The activation value after the first layer for $v^{th}$ hidden SVM is therefore defined as:

$$\vec{h}^{(1)}(v) = \sum_{i=1}^{M} \vec{\alpha}_{vi} \vec{y}_i K_v(\vec{x}_{vs}, \vec{x}_{vi}), \qquad (15)$$

where $v$ represents the $v^{th}$ SVM in a specific hidden layer.

The activation values within the hidden layers are defined in a general form as:

$$\vec{h}^{(l)}(v) = \sum_{i=1}^{M} \vec{\alpha}_{vi} \vec{y}_i K_v(\vec{x}_{vs}, \vec{x}_{vi}) \qquad (16)$$

Here, $\vec{h}^{(l)}(v) \in \mathbb{R}^v$ is the $v$-dimensional $v^{th}$ element of the $l^{th}$ layer. The bias $b$ has been ignored in the middle layers as this is just bias for the classification and does not affect the fundamental distribution of the data. The classification of an unknown input $\vec{x}$ at the final layer with multiclass SVM is done with the following formulation:

$$y(\vec{x}) = \sum_{s=1}^{S} \vec{\alpha}_s \vec{y}_s K_s\left(\vec{x}_s, \ddot{O}(\vec{x})\right) + b, \qquad (17)$$

where $S$ is the number of support vectors at the last layer, $\vec{x}_s$ is the $s^{th}$ support vector, and $\ddot{O}(\vec{x})$ is the transformed feature vector of the input $\vec{x}$ by the hidden layers. We can now obtain SVM parameters either by solving the quadratic programming problem formulation or solving a system of linear equations (also known as least squares SVM).

In section 2, we discussed how to formulate quantum binary and multiclass SVM. We use this formulation to design the framework for quantum deep SVM.

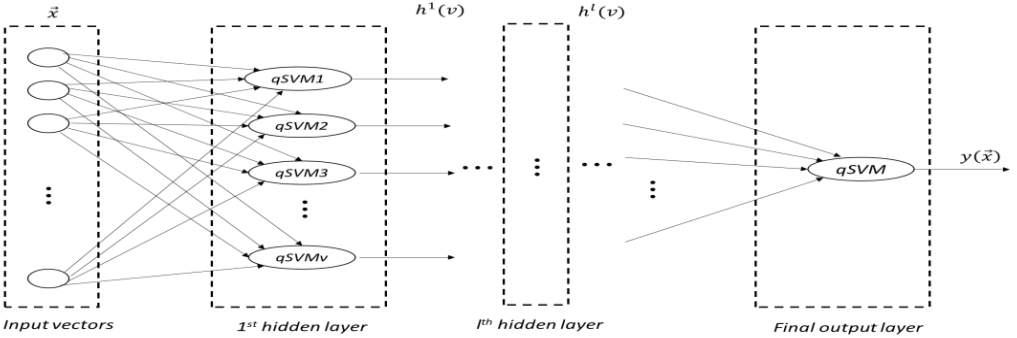

Fig.2: quantum Deep SVM architecture

Referring to the above **Fig.2**, in quantum setting, for each hidden quantum multiclass least squares SVM $v$ we determine the parameters $(|b_v^l, \vec{\alpha}_v^l\rangle)_r$ for a specific hidden layer $l$, where subscript $r$ represents the $r^{th}$ quantum binary classfier in multiclass setting **[12]**, $r = 1, 2, \ldots, \frac{g(g-1)}{2}$, and

$$(|b_v^l, \vec{\alpha}_v^l\rangle)_r = \left(\left(\hat{F}_v^l\right)^{-1} |\vec{y}\rangle\right)_r. \qquad (18)$$

We proceed now as discussed in the section 2 to solve the above equation. For simplicity we are presenting the solution for a single quantum binary SVM, which results

$$|b_v^l, \vec{\alpha}_v^l\rangle = \frac{1}{\sqrt{b^2 + \sum_{i=1}^{M_{max}}(\alpha_{vi}^l)^2}} \left(b|0\rangle + \sum_{i=1}^{M_{max}} \alpha_{vi}^l |i\rangle\right). \qquad (19)$$

Where $M_{max} = max\{M_{i,j \in r \,\&\, i \neq j}\} \in M$, and $M_{i,j}$ is the input data set for the classifier whose date set contains only the data of $i^{th}$ & $j^{th}$ classes. $M$ is the total number of training inputs.

In a similar way, we can simply extend it to define $r$ quantum binary SVM classifiers and apply quantum all-pair algorithm to formulate multiclass case **[12]**. Upon training a specific quantum SVM at a layer $l$, the outputs of the $(v^l)^{th}$ quantum multiclass SVMs are used as training input vectors for the quantum multiclass SVMs $v^{l+1}$ in the next hidden layer $l+1$. The dimension of an input vector is $v$ in this case. In quantum SVM, the output activation function $|h_v^l\rangle$ of a quantum multiclass SVM $v^l$ in $l^{th}$ layer is a function of $b$ and $\vec{\alpha}_v^l$. Where $|h_v^l\rangle = concate(|h_v^l\rangle_1, |h_v^l\rangle_1, \ldots, |h_v^l\rangle_r)$, $|h_v^l\rangle_r$ is the activated value of the $r^{th}$ quantum binary classifier of the $v^{th}$ quantum multiclass SVM in the $l^{th}$ layer, and $\vec{\alpha}_v^l = concate((\vec{\alpha}_v^l)_1, (\vec{\alpha}_v^l)_2, \ldots, (\vec{\alpha}_v^l)_r)$.

We ignore $b$ as this is just a bias and does not affect the data distribution in the hidden layers. Therefore, $|h_v^l\rangle$ is now a function of $\vec{\alpha}_v^l$ only within the hidden layers. The second last layer's outputs (aka activation values from the second last layer) is now the input vectors for the last layer quantum multiclass SVM. The last layer parameter formulation is now defined as:

$$(|b^{last\_layer}, \vec{\alpha}^{last\_layer}\rangle)_r = \left(\left(\hat{F}^{last\_layer}\right)^{-1}|\vec{y}\rangle\right)_r ; r = 1, 2, \ldots, \frac{g(g-1)}{2}, \qquad (20)$$

Where $r$ represents the case of $r^{th}$ classifier,

$$\hat{F}^{last\_layer} = (K^{last\_layer} + J + (\eta^{last\_layer})^{-1}I)/trF^{last\_layer},$$

$$\hat{K}^{last\_layer} = \frac{K^{last\_layer}}{trK^{last\_layer}} = 1/trK^{last\_layer} \sum_{i,j=1}^{M_{max}} \langle \ddot{O}(\vec{x}_i)|\ddot{O}(\vec{x}_j)\rangle |\ddot{O}(\vec{x}_i)||\ddot{O}(\vec{x}_i)||i\rangle\langle j|, \quad (21)$$

and $\ddot{O}(\vec{x}_i) = h_v^{second\_last\_layer}$ is the transformed input vector $|\vec{x}_i\rangle$ by multiple hidden layers (the activation value from the second last layer).

The parameters of the final trained quantum deep SVM is determined by,

$$|b^{last\_layer}, \vec{\alpha}^{last\_layer}\rangle = \frac{1}{\sqrt{(b^{last\_layer})^2 + \sum_{i=1}^{M_{max}}\left(\alpha_i^{last\_layer}\right)^2}} \left(b^{last\_layer}|0\rangle + \sum_{i=1}^{M_{max}} \alpha_i^{last\_layer}|i\rangle\right).$$

$$(22)$$

Where $\vec{\alpha}_v^{last\_layer} = concate\left((\vec{\alpha}_v^{last\_layer})_1, (\vec{\alpha}_v^{last\_layer})_2, \ldots, (\vec{\alpha}_v^{last\_layer})_r\right)$. We have trained the quantum deep SVM, now for classifying an unknown vector $|\vec{x}\rangle$, we perform a swap test between $(|b^{last\_layer}, \vec{\alpha}^{last\_layer}\rangle)_{r; r=1,2,\ldots,\frac{g(g-1)}{2}}$ and $|\vec{x}\rangle$ and based on the success probabilities and by using quantum all pair algorithm we identify the predicted class. We prepared a training data oracle from equation **(22)**:

$$(|\tilde{T}\rangle)_{r; r=1,2,\ldots,\frac{g(g-1)}{2}} = \frac{1}{\sqrt{(b^{last\_layer})^2 + \sum_{i=1}^{M_{max}}\left(\alpha_i^{last\_layer}\right)^2 |\vec{x}_i|^2}} \left(b^{last\_layer}|0\rangle|0\rangle + \sum_{i=1}^{M_{max}}\left(\alpha_i^{last\_layer}\right)^2 |\vec{x}_i||i\rangle|\vec{x}_i\rangle\right)$$

$$(23)$$

We also construct a query state,

$$|\tilde{x}\rangle = \frac{1}{\sqrt{M_{max}|\vec{x}|^2+1}}\left(|0\rangle|0\rangle + \sum_{i=1}^{M_{max}}|\vec{x}||i\rangle|\vec{x}\rangle\right) \tag{24}$$

Using the ancillas, we further construct the following states,

$$(|\Psi\rangle)_{r;\, r=1,2,\ldots,\frac{g(g-1)}{2}} = \frac{1}{\sqrt{2}}\left(|0\rangle(|\tilde{T}\rangle)_{r;\, r=1,2,\ldots,\frac{g(g-1)}{2}} + |1\rangle|\tilde{x}\rangle\right), \tag{25}$$

and, measure the ancillas in the following state,

$$|\Phi\rangle = \frac{1}{\sqrt{2}}(|0\rangle - |1\rangle), \tag{26}$$

Upon measurement, the success probabilities are determined by $(P)_{r;\, r=1,2,\ldots,\frac{g(g-1)}{2}} = |(\langle\Psi|)_{r;\, r=1,2,\ldots,\frac{g(g-1)}{2}}|\Phi\rangle|^2 = \frac{1}{2}\left(1 - (\langle\tilde{T}|)_{r;\, r=1,2,\ldots,\frac{g(g-1)}{2}}|\tilde{x}\rangle\right)$. If $(P)_{r;\, r=1,2,\ldots,\frac{g(g-1)}{2}} < 1/2$ we classify the input vector $|\vec{x}\rangle$ as $+1$, otherwise $-1$. And, then by using the quatum all pair algorithm **[12]**, we predict the class.

## 5 QUANTUM CLUSTERING APPROACH WITH QUANTUM DEEP SVM

We have formulated a quantum deep SVM in the previous section and now we discuss a deep quantum clustering framework based on the quantum deep SVM. Recently, in **[3]**, a research group demonstrated a very interesting deep clustering framework with pre-trained convolutional neural networks and *K*-Means clustering in the classical domain. The framework contains a pre-trained deep convolutional neural network, which extracts the features from the unlabeled training vectors (in this case the input vectors are images). These features are then fed into *K*-Means clustering algorithms and the cluster outputs are considered as the pseudo labels for the images. We then trained the deep CNN end to end with the training vectors and its pseudo labels $\left\{y_i^{(pseudo)} \in \{0,1\}^g\right\}$ using gradient descent algorithm and optimize the total combined loss of the model (loss of the deep CNN plus the loss of the *K*-Means). Here, $g$ is the number of pseudo-classes. We repeat multiple epochs of training until we reach an acceptable clustering result. The objective is to optimize the following cost function:

$$min_{\theta,\overline{w}} \frac{1}{M}\sum_{i=1}^{M} \Gamma\left(Z(f_\theta(\vec{x}_i)), y_i^{(pseudo)}\right), \tag{27}$$

where $Z$ is a parameterized classifier, $y_i^{(pseudo)}$ represents the pseudo label for the $i^{th}$ input vector $\vec{x}_i$ produced by a standard clustering algorithm *K*-Means (clustering algorithm choice is just an arbitrary here and can be tried with any other standard clustering algorithm), and $\Gamma$ is the negative log-softmax function. In this strategy, the K-Means takes the input feature vectors $f_\theta(\vec{x}_i)$ produced by the pre-trained deep CNN and cluster them into distinct $g$ groups by solving the following problem:

$$min_{C \in \mathbb{R}^{d \times g}} \frac{1}{M} \sum_{i=1}^{M} min_{y_i^{(pseudo)} \in \{0,1\}^g} \left\| f_\theta(\vec{x}_i) - B y_i^{(pseudo)} \right\|_2^2 \quad (28)$$

where B is the $d \times g$ centroid matrix, and $\left(y_i^{(pseudo)}\right)^T (1_g) = 1$.

Solving the equation **(28)**, we get the optimal cluster assignments. These cluster assignments are used as pseudo labels for training input vectors. This concludes that the formulation of deep clustering alternates between clustering the features to generate the pseudo labels using equation **(28)**. The optimized labels update the deep neural network's parameters by predicting these pseudo labels using the equation **(27)**. This helps in generating optimized clustering assignments at the end after $E$ number of training epochs including both the deep neural network and K-Means clustering loss function optimization.

In quantum setting, we keep the classical pre-trained CNN but we replaced the softmax function in the last layer with a quantum deep multiclass SVM. We also replace the classical K-Means clustering algorithm with a quantum version of the $K$-Means clustering algorithm. We optimize the cost function of the CNN by using the classical back propagation algorithm with gradient descent algorithm. In quantum framework, we only optimize the deep CNN parameters and avoid optimizing the quantum deep SVM parameters during the back propagation.

By introducing quantum deep multiclass SVM, at last, the layer, we get performance improvement. In addition, the quantum version of $K$-Means improves the performance of the framework further. **Fig.3** demonstrates the proposed quantum deep SVM clustering approach.

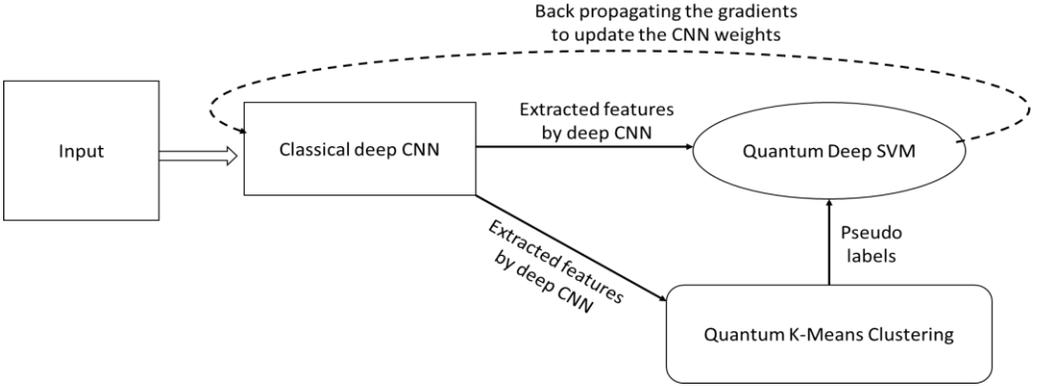

Fig.3: Quantum deep clustering architecture

In quantum version the, we target to optimize the following cost function using the back propagating the gradients in deep CNN:

$$L(\vec{w}) = min_{\theta,\vec{w}} \frac{1}{M} \sum_{i=1}^{M} (\vec{w})^T (\vec{w}) + C \sum_{i=1}^{M} max\left(1 - (\vec{w})^T f_\theta(\vec{x}_i) y_i^{(pseudo)}, 0\right)^2, \quad (29)$$

The lower layer weights of the deep CNN are learned by back propagating the gradients from the top quantum deep SVM. For the purpose, we differentiate the above objective function defined in equation **(29)** with respect to an activation of the penultimate layer. For the demonstration purpose, the $f_\theta(\vec{x}_i)$ is replaced with an activation function $h_i$, we get:

$$\frac{\partial L(\vec{w})}{\partial h_i} = -2Cy_i^{(pseudo)}\vec{w}\left(max\left(1 - w^T h_i y_i^{(pseudo)}, 0\right)\right) \tag{30}$$

Please note that we are not updating the quantum deep SVM parameters during back propagation. These parameters are updated by using the method discussed in **section 4**.

The features $f_\theta(\vec{x}_i)$ produced by the deep CNN is feed into the quantum $K$-Means clustering to produce optimized clusters as well as the pseudo labels $y_i^{(pseudo)}$, $\cong$ the output cluster assignments with $c_i$, for the quantum deep SVM classifier. In our settings, outputs of the quantum $K$-Means clustering algorithm are in superposition by using an adiabatic algorithm and represented by the following quantum state:

$$|\varsigma\rangle = \left(\frac{1}{\sqrt{M}}\right)\sum_j |c_j\rangle|j\rangle = \left(\frac{1}{\sqrt{M}}\right)\sum_c \sum_{j \in c} |c\rangle |j\rangle \tag{31}$$

We select $K^{(clustering)}$ vectors with labels $i_c$ as initial seeds for each of the clusters. By using the adiabatic theorem, we now assign the rest of the vectors to these clusters, $K^{(clustering)}$ is the number of clusters. We began with the state

$$\frac{1}{\sqrt{MK^{(clustering)}}} \sum_{c'j} |c'\rangle|j\rangle \left(\frac{1}{\sqrt{K^{(clustering)}}} \sum_c |c\rangle|i_c\rangle\right)^{\otimes D}$$
$$(32)$$

where the $D$ copies of the state $\frac{1}{\sqrt{K^{(clustering)}}}\sum_c |c\rangle|i_c\rangle$ combined with the distance evaluation technique given in **[6]** allow us to evaluate the distance $\left|f_\theta(\vec{x}) - f_\theta\left(\vec{x}_{i_{c'}}\right)\right|^2$ in the $c'j$ component in the superposition. The result is the initial clustering state

$$|\psi_1\rangle = \frac{1}{\sqrt{M}}\sum_{c,j \in c}|c\rangle|j\rangle \tag{33}$$

where the state $|j\rangle$ is associated with the cluster $c$ with the closet seed vector $i_c$. We now construct the following cluster state by constructing $D$ copies of the above state $|\psi_1\rangle$:

$$|\phi_1^c\rangle = \frac{1}{\sqrt{M_c}}\sum_{j \in c}|j\rangle. \tag{34}$$

We estimate the number of states $M_c$ in the cluster $c$. Now with state $|\phi_1^c\rangle$ and $D$ copies of the state $|\psi_1\rangle$, we evaluate the average distance between $f_\theta(\vec{x}_j)$ and the mean of the cluster c:

$$\left|f_\theta(\vec{x}_j) - \left(\frac{1}{M_c}\right)\sum_{k \in c'} f_\theta(\vec{x}_k)\right|^2 = \left|f_\theta(\vec{x}_j) - \overline{f_\theta(\vec{x}_c)}\right|^2 \tag{35}$$

We then apply a phase to each component $|c'\rangle|j\rangle$ of the superposition with the following Hamiltonian:

$$H_f = \sum_{c',j}|f_\theta(\vec{x}_j) - \overline{f_\theta(\vec{x}_{c'})}|^2 |c'\rangle\langle c'|\otimes|j\rangle\langle j|\otimes I^{\otimes D} \tag{36}$$

Now, with an initial Hamiltonian $H_{initial} = 1 - |\phi\rangle\langle\phi|$, where $|\phi\rangle = \frac{1}{\sqrt{K}}\sum_{c'}|c'\rangle$ is the superposition of cluster centroids, we perform the adiabatic evolution with the state

$$\frac{1}{\sqrt{MK^{(clustering)}}}\sum_{c',j}|c'\rangle|j\rangle|\psi_1\rangle^{\otimes D} \tag{37}$$

We finally, get the following final state

$$\left(\frac{1}{\sqrt{M}}\sum_{c',j\in c'}|c'\rangle|j\rangle\right)|\psi_1\rangle^{\otimes D} = |\psi_2\rangle|\psi_1\rangle^{\otimes D} \tag{38}$$

We now need to repeat this $D$ times to generate $D$ copies of the restructured state $|\psi_2\rangle$. We repeat this cluster assignment at each step in successive iterations until clustered quantum state $|\varsigma\rangle$ contains the final optimized clusters in quantum superposition. These clusters assignments information (label information in actual, $y_i^{(pseudo)}$) along with features $f_\theta(\vec{x}_j)$ are then again used to generate the input quantum states in superposition for the quantum deep classifier with the help of quantum random access memory (QRAM) **[18]**.

## 6   ERROR & COMPUTATIONAL COMPLEXITY ANALYSIS

At first, we break down the discussion on the error & computational complexity analysis into three sections – deep CNN, quantum deep SVM, and quantum $K$-Means clustering. In the end, we investigate & conclude the overall computational complexity of the proposed quantum clustering framework, and its speed up gain against the probable classical framework. In our proposed quantum clustering framework we are open to using any pre-trained deep CNN, so we are exploring the computational complexity in general sense and can be modified based on the specific types of pre-trained deep CNN. The deep CNN associates with two functional approaches- forward propagation and backward propagation **[19]**. During forward propagation, we formulate the activation $g(z^{(l)})$ function as

$$g(z^{(l)}) = g(\vec{w}_{(CNN)}^{(l)} a^{(l-1)}); \ z^{(l)} \in \mathbb{R}^{1\times(m|\vec{w}_{(CNN)}^{(l)} \in \mathbb{R}^{m\times n})} \tag{39}$$

where $a^{(l-1)}$ is the activated value of the $(l-1)^{th}$ layer, $\vec{w}_{(CNN)}^{(l)}$ is the weight parameters of the CNN in the $l^{th}$ layer.

We observe that there is a matrix multiplication operation and an activation function computation in each layer in the network. In general, a matrix multiplication has a run time complexity in the order of $O(n^3)$. But, in the case of element-wise activation function $g(z^{(l)})$, the run time is in the order of $O(n)$. Where $n$ is the cardinality of the row (or column) vector of a square matrix. We define the weights of the network in $l^{th}$ layer as:

$$\vec{w}^l_{CNN} = \begin{cases} \mathbb{R}^{n^{(l)} \times 1} & if\ l = 0 \\ \mathbb{R}^{n^{(l)} \times n^{(l-1)}} & if\ l > 0 \end{cases} \tag{40}$$

where $n^{(l)}$ is the number of neurons in $l^{th}$ layer.

The total number of multiplications in the network is there:

$$n_{mul} = \sum_{l=2}^{L}\left(n^{(l)} n^{(l-1)} n^{(l-2)}\right) + \left(n^{(1)} n^{(0)} 1\right), \tag{41}$$

where $L$ is the number of layers in the network. Also, the number of times we apply the activation functions:

$$n_{act} = \sum_{l=1}^{L}\left(n^{(l)}\right). \tag{42}$$

The total run time complexity in the case of forward propagation is there:

$$T_{forward} = n_{mul} + n_{act} = \sum_{l=2}^{L}\left(n^{(l)} n^{(l-1)} n^{(l-2)}\right) + \left(n^{(1)} n^{(0)} 1\right) + \sum_{l=1}^{L}\left(n^{(l)}\right). \tag{43}$$

$$T_{forward} \simeq L \cdot (N^3) + L \cdot N \simeq O(L \cdot N^3) + O(L \cdot N) \approx O(LN^3), \tag{44}$$

where $N$ is the dimension of the input feature vector and assuming $n^{(l)} \subseteq N$.

During the back propagation, we first calculate the total error at the last layer, next we use the intermediate layer's delta errors to calculate (using the chain rule) the error of all the previous layers and at the last, and we calculate the derivative of the error function with respect to each weight in the network. (Gradient descent algorithm used these weights for required weight updates). The total time required in back propagation is the time taken by the gradient descent algorithm considering the time of weights finding $(T_{back\_weights})$ and delta error calculation in the layers $(T_{back\_error})$. We have,

$$T_{back\_error} = \begin{cases} \nabla_h^{(L)} \odot g'(z^{(L)}), & if\ l = L \\ \vec{w}^{(l+1)^T} \delta^{(l+1)} \odot g'(z^{(l)}), & otherwise \end{cases}, \tag{45}$$

$$\tag{45}$$

where $\delta^{(l)} = \frac{\partial L(\vec{w})}{\partial h_i^{(l)}}$ (from the equation (30)), $\partial h_i^{(l)}$ is the activated value at $l^{th}$ layer associated with $i^{th}$ weight. $\nabla_h^{(L)}$ is the vector gradient. $z^{(L)}$ is the output value of a neuron in the $l^{th}$ layer before applying the activation function.

$$T_{back\_error} = O(n^2 + n^3(L-1)) \approx O(Ln^3), \tag{46}$$

where $n$ is the number of neurons in a layer and $L$ is the total number of layers in the network.

Also,

$$T_{back\_weights} = T_{back\_error} + n^3 \approx O(Ln^3 + n^3) \approx O(Ln^3) \tag{47}$$

So, the overall run time of the back propagation is:

$$T_{back} = Gr \cdot O(Ln^3) \approx O((Gr)Ln^3) \approx O((Gr)LN^3), \tag{48}$$

where $Gr$ is the number of iterations the Gradient descent takes to optimize the weights in the network, and $N$ is the dimension of the input feature vector and assuming $n^{(l)} \subseteq N$.

Therefore, CNN takes:

$$T_{forward} + T_{back} + T_{conv},$$
$$= O(Ln^3) + O((Gr)LN^3) + T_{conv} = O((Gr)LN^3) + T_{conv}, \tag{49}$$

where $T_{conv}$ is the time to calculate convolutional operations. We are not explaining the calculating of the $T_{conv}$ as this depends on the pre-trained network architecture. With an error factor $\epsilon_{gd}$, Gradient descent possess a convergence rate of $O\left(\frac{1}{\epsilon_{gd}}\right)$ for a convex function.

During the quantum setting of SVM, the input vector $|\vec{x}_i\rangle$ is in the superposition:

$$|\vec{x}_i\rangle = \frac{1}{|\vec{x}_i|}\sum_{j=1}^{N}(\vec{x}_i)_j |j\rangle, \tag{50}$$

We construct these states by using quantum RAM **[18]**, which uses $O(MN)$ resources but takes only $O(logMN)$ to access them. The quantum SVM has sped up to gain in two sections-in Kernel matrix calculation & during the training. Quantum matrix inversion is an important process in quantum SVM, which performs a quantum PCA **[10]**. $\hat{F}$ contains kernel matrix $\hat{K}$ and offset parameter $b$. If we ignore the offset parameter (assuming its value is negligible), the discussion on computation complexity of the matrix $\hat{F}$ reduces to the matrix inversion of $\hat{K}$. We define the conditional number, $\kappa$, as the ratio of the largest eigenvalue to the smallest eigenvalue. Referring **[9]**, the eigenvalue of $\hat{F}$ is defined as $O\left(\frac{1}{M}\right) \leq \lambda_{\hat{F}} < 1$. Therefore, $\kappa = O(M)$. Solving such eigenvalue is an exponentially costly **[8** Using phase estimation and filtering process in quantum setting **[8]**, we find the kernel matrix's principle. We need to define a constant $\epsilon_K$ such that its value bounds the lower limit of the eigenvalues under consideration. We then apply the filtering procedure **[8]** using the kernel matrix condition value $\frac{1}{\epsilon_K}$. Let, $T = \frac{t_0}{\Delta t}$ be the phase estimation run-time steps, where $t_0$ is the time to evaluate the error in the phase estimation, and $\Delta t$ is the time interval. The propagator $e^{-i\mathcal{L}_R \Delta t}(I)$ in equation **(13)** during the phase estimation subjects to the error $O(\Delta t^2 ||\hat{K}) \approx O(\Delta t^2 ||\hat{F})$. When we take the powers of the propagation under consideration, we observed the error of maximally

$\epsilon = O(T\Delta t^2) = O\left(T\left(\frac{t_0}{T}\right)^2\right) = O(t_0^2 T^{-1}) \to T = O(t_0^2 \epsilon^{-1})$. At this point, the overall run time is $O(t_0^2 \epsilon^{-1} \log MN)$, where the kernel matrix preparation takes $O(\log MN)$ run time in the quantum setting. When we consider the relative error due to phase estimation **[8]**, the run time becomes $O\left(\left(\frac{1}{\epsilon_K \epsilon}\right)^2 \epsilon^{-1} \log MN\right) \approx O(\epsilon_K^{-2} \epsilon^{-3} \log MN)$. Therefore, the overall run rime computational complexity of a quantum binary SVM scales on $O(\log MN)$. In case of multiclass classification with quantum all-pair approach **[12]**, the run time becomes $O(g^2 \log M_{max} N) + O(g^{3/2} \log(\delta^{-1})\varepsilon^{-1}) + O(\log g)$, where $\varepsilon$ is the measurement error, $\delta$ is a parameter, and $M_{max} < (and \in) M$ (for detail please refer our paper **[12]**).

In the proposed quantum deep clustering, the total runtime of the deep quantum SVMs will be the sum of the total run time of the quantum multiclass SVMs at hidden layers and the last single quantum multiclass SVM, i.e.

$$[O((v+1)lg^2 \log M_{max} N) + O((v+1)lg^{3/2} \log(\delta^{-1})\varepsilon^{-1}) + O((v+1)l \log g)]. \quad (51)$$

where $v$ is the total number of quantum multiclass SVMs and $l$ is the number of layers in deep SVM.

The quantum $K$-Means clustering's input and output are defined in quantum states as we discussed in early sections. These give us an exponential speed-up gain advantage. The algorithm implementable with $O(\epsilon_{KMeans} K^{(clustering)} \log K^{(clustering)} MN_{f_\theta(\vec{x}_i)})$ run time computational complexity, where $\epsilon_{KMeans}$ is an accuracy factor. In fact, if the clusters are relatively well separated, the run time even becomes $O(\epsilon_{KMeans} \log K^{(clustering)} MN_{f_\theta(\vec{x}_i)})$. Where the $N_{f_\theta(\vec{x}_i)}$ is the dimension of the input feature vector for the quantum $K$-Means clustering algorithm, and $K^{(clustering)}$ is the number of clusters.

We now compare the overall run time complexity of our proposed quantum deep clustering algorithm against the possible classical implementation. Please refer to **Table.1**:

Table.1: Run time computational complexity analysis

| **Run time contributors** | **Run time with the classical approach** | **Run time with the quantum approach** | **Remark** |
|---|---|---|---|
| Convolution Neural Network | $T_{C1} = O((Gr)LN^3) + T_{conv}$ | $T_{q1} = O((Gr)LN^3) + T_{conv}$ | No advantages in run time. |
| Multiclass SVMs at the hidden and last layers in the deep SVM | $T_{C2} = O((v+1)lg^2 M^3)$ | $T_{q2} = [O((v+1)lg^2 \log M_{max} N) + O((v+1)lg^{3/2} \log(\delta^{-1})\varepsilon^{-1}) + O((v+1)l \log g)]$ | Clearly significant quantum advantages. Approximately exponential speed-up gain. |
| $K$-Means clustering | $T_{C3} = O(K^{(clustering)} MN_{f_\theta(\vec{x}_i)})$ | $T_{q3} = O(\log K^{(clustering)} MN_{f_\theta(\vec{x}_i)})$ | Clearly significant quantum advantages. Approximately exponential speed-up gain. |
| Overall run time | $T_{C1} + T_{C2} + T_{C3}$ | $T_{q1} + T_{q2} + T_{q3}$ | Significantly speed up gain in quantum setting as compared to the classical setting. |

# 7   CONCLUSIONS

In this paper, we presented an innovative quantum deep clustering framework and our theoretical research have shown that the proposed quantum deep clustering framework is approximately exponentially faster than the classical possible implementation at most of the sections in the framework, and overall demonstrated significant speed up advantages against the classical implementation. In our work, we first constructed a quantum deep SVM formulation and then extended the formulation to develop a quantum deep clustering framework using a classical deep convolutional neural network (CNN) (choice of the CNN architecture depends on the specific requirement and has been generic in this research for broader flexibility of using pre-trained CNNs) and a quantum $K$-Means clustering. Our research works also demonstrates a first of its kind of quantum-classical hybrid machine learning formulation. Our research investigations open and expend a new interesting research horizon towards the quantum-classical machine learning origination on practical ground.